# Facial Recognition Technology and Human Raters Can Predict Political Orientation From Images of Expressionless Faces Even When Controlling for Demographics and Self-Presentation

Michal Kosinski, Poruz Khambatta, and Yilun Wang
Graduate School of Business, Knight Management Center, Stanford University

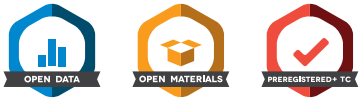

Carefully standardized facial images of 591 participants were taken in the laboratory while controlling for self-presentation, facial expression, head orientation, and image properties. They were presented to human raters and a facial recognition algorithm: both humans ($r = .21$) and the algorithm ($r = .22$) could predict participants' scores on a political orientation scale (Cronbach's $\alpha = .94$) decorrelated with age, gender, and ethnicity. These effects are on par with how well job interviews predict job success, or alcohol drives aggressiveness. The algorithm's predictive accuracy was even higher ($r = .31$) when it leveraged information on participants' age, gender, and ethnicity. Moreover, the associations between facial appearance and political orientation seem to generalize beyond our sample: The predictive model derived from standardized images (while controlling for age, gender, and ethnicity) could predict political orientation ($r \approx .13$) from naturalistic images of 3,401 politicians from the United States, the United Kingdom, and Canada. The analysis of facial features associated with political orientation revealed that conservatives tended to have larger lower faces. The predictability of political orientation from standardized images has critical implications for privacy, the regulation of facial recognition technology, and understanding the origins and consequences of political orientation.

*Public Significance Statement*
We demonstrate that political orientation can be predicted from neutral facial images by both humans and algorithms, even when factors like age, gender, and ethnicity are accounted for. This indicates a connection between political leanings and inherent facial characteristics, which are largely beyond an individual's control. Our findings underscore the urgency for scholars, the public, and policymakers to recognize and address the potential risks of facial recognition technology to personal privacy.

*Supplemental materials:* https://doi.org/10.1037/amp0001295.supp

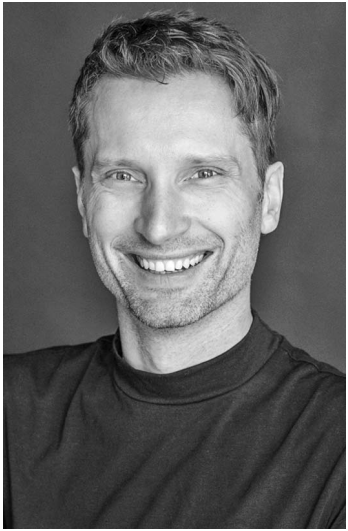

**Michal Kosinski**

The widespread use of facial recognition technology poses serious challenges to privacy and civil liberties. Policymakers, scholars, and the public are increasingly alarmed by how easy it is to identify individuals in images, as well as to track their location and social interactions (Santow, 2020). However, pervasive surveillance is not the only problem posed by facial recognition. Apart from identifying individuals, facial recognition algorithms can also identify their personal attributes. Just as humans can, facial recognition algorithms can also predict individuals' emotions, age, gender, and ethnicity (Mollahosseini et al., 2019; Ranjan et al., 2018). Even more worrisome, these algorithms can identify personal attributes that were, thus far, widely considered to be unrecognizable from faces (Todorov et al., 2015). Patents filed by organizations ranging from startups to Xerox (Bart & Biswas, 2014; Wilf et al., 2012)—as well as a recent flurry of scientific articles—show that facial recognition can accurately infer sensitive traits such as political orientation (Joo et al., 2015; Kosinski, 2021; Rasmussen et al., 2023), personality (Kachur et al., 2020; Moreno-Armendariz et al., 2020; Segalin et al., 2017; Xu et al., 2021), and sexual orientation (Leuner, 2019; D. Wang, 2022; Y. Wang & Kosinski, 2018).

What remains unclear is the extent to which face-based predictions of personal attributes are enabled by stable facial features over which people have little control (e.g., facial attractiveness and morphology) versus factors that people can—to some extent—influence (e.g., facial expression and hairstyle; Todorov et al., 2015). Here, we explore this issue in the context of one of the major psychological traits: political orientation. We note that political orientation is related to but not synonymous with party affiliation. The latter is often shaped by regional, historical, cultural, and specific policy-related factors unique to a particular nation. It is a tangible association, typically demarcated by voting behaviors, and can vary widely across countries. On the other hand, political orientation is a more universal psychological construct that refers to an individual's fundamental beliefs, values, and tendencies, typically placed on a conservative to liberal spectrum. While this orientation can influence party affiliation, it transcends national or group boundaries (Jost et al., 2009).

Past research indicates that political orientation can be inferred from facial images by both humans (Jahoda, 1954; Rule & Ambady, 2010; Samochowiec et al., 2010) and facial recognition algorithms (Joo et al., 2015; Kosinski, 2021; Rasmussen et al., 2023). Human judgments—while of relatively low accuracy (Rule & Ambady, 2010)—are instant, instinctive, consistent across raters, and reliably better than random guessing (Tskhay & Rule, 2013). Facial recognition algorithms' accuracy tends to be higher; a single facial image is as revealing of political orientation as one's responses to a 100-item personality questionnaire (Kosinski, 2021).

However, the design of past studies hinders our understanding of whether stable facial features or other factors, such as self-presentation, are the primary indicators of political orientation. Many previous studies relied on self-selected facial images (e.g., social media profile pictures) that contained confounding factors potentially related to political orientation. Such variables include self-presentation (e.g., makeup, facial hair style, and head orientation), facial expression, and image properties such as resolution and sharpness (D. Wang, 2022; Y. Wang & Kosinski, 2018). While these variables may hint at political orientation, they could also mask the associations between political orientation and stable facial features. For instance, facial hairstyle, head orientation, or makeup may correlate with political orientation, but they can also obscure the associations between political orientation and facial morphology.

Additionally, a significant portion of past research used images of elected politicians. This is problematic, as the differences between the faces of liberal and conservative politicians do not necessarily imply that the faces of their respective electorates also differ. Instead, they could reflect the left–right differences in voters' preferences for candidates' facial appearance (Olivola et al., 2012).

Due to the limitations of past studies, many scholars have concluded that the predictability of political orientation (and other traits) is driven by self-presentation, demographics, or image properties rather than by stable facial features (Todorov et al., 2015). However, such a conclusion seems premature. The prior studies' shortcomings render them inconclusive in this regard: The predictability might stem solely from stable features or, more plausibly, from a combination of factors, including stable facial features, self-presentation, demographics, and image properties. Moreover, rejecting the existence of links between stable facial features and political orientation contradicts many well-known mechanisms that imply the existence of such

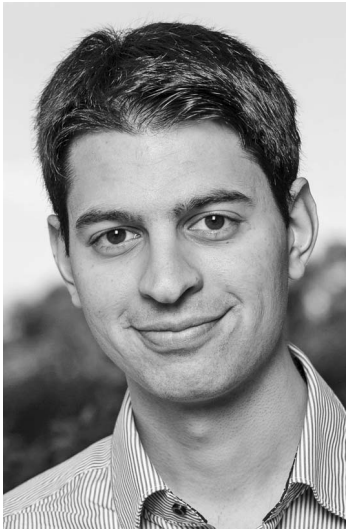

**Poruz Khambatta**

links. These mechanisms can be grouped into three causal pathways (see Figure 1).

First, facial appearance can shape psychological traits (*face → mind* pathway). People largely agree when judging political orientation from faces (Todorov et al., 2015). Regardless of whether such judgments are accurate, the self-fulfilling prophecy effect (Merton, 1936) postulates that people perceived as having a particular attribute are treated accordingly; internalize such attributions; and, over time, may engage in behaviors consistent with others' perceptions (Slepian & Ames, 2016). For example, people with larger jaws, often perceived as more socially dominant (a trait associated with political conservatism), might over time become more so (Wilson & Sibley, 2013; Windhager et al., 2011). Moreover, face-based perceptions influence consequential outcomes such as the length of prison sentences, occupational success, educational attainments, the chances of winning an election, income, and status (Ballew & Todorov, 2007; Eberhardt et al., 2006; Todorov et al., 2005; Zebrowitz & Montepare, 2008). Those outcomes, in turn, shape people's political attitudes (Zebrowitz & Montepare, 2008). Becoming wealthier, for example, shifts people toward political conservatism (Peterson, 2016).

Second, latent factors shape both psychological traits and facial appearance (*face ← factor → mind* pathway). Those include socioeconomic status, environmental and developmental conditions, hormones, and genes. Twin studies, for example, have found that genes are responsible for over 50% of the variation in both facial features (Richmond et al., 2018) and political orientation (Alford et al., 2005). Furthermore, prenatal exposure to nicotine and alcohol affects facial morphology (Richmond et al., 2018) and cognitive ability, which is associated with political orientation (Onraet et al., 2015).

Third, psychological traits can shape facial appearance (*mind → face* pathway, or the Dorian Gray effect; Zebrowitz, 2018). While we tend to think of facial features as relatively fixed, they are shaped by factors such as facial care, diet, substance use, physical health, injuries, exposure to sunlight, harsh environmental conditions, or emotional states (Richmond et al., 2018). Exposure to such face-altering factors, in turn, is associated with psychological traits. Liberals, for example, tend to smile more intensely and genuinely (Wojcik et al., 2015), which, over time, leaves traces in wrinkle patterns (Piérard et al., 2003). Conservatives tend to be more self-disciplined and are thus healthier, consume less alcohol and tobacco, and have a better diet (Chan, 2019; Subramanian & Perkins, 2010), altering their facial fat distribution and skin health (Richmond et al., 2018).

Naturally, some of those (and other similar) mechanisms may be misconceived or have a negligible effect. Yet, given their number and diversity, it is likely that there are some links between stable facial features and political orientation. This work aims to test this possibility while attempting to address some of the limitations of previous studies. Study 1 shows that facial recognition algorithms can accurately predict political orientation from carefully standardized images of neutral faces while controlling for demographics, self-presentation, facial expression, and image properties. (We use a sample of nonpoliticians to control for the potential nonrepresentativeness of politicians' faces.) Study 2 presents the same task to human raters who achieved comparable performance. Study 3 validates the model trained in Study 1 on a very different sample: profile images of politicians from the United States, the United Kingdom, and Canada. The predictability holds across countries, age groups, genders, Black and White ethnicities, and facial expressions. This suggests that the links between

**Figure 1**

*Three Causal Pathways Linking Facial Features and Political Orientation*

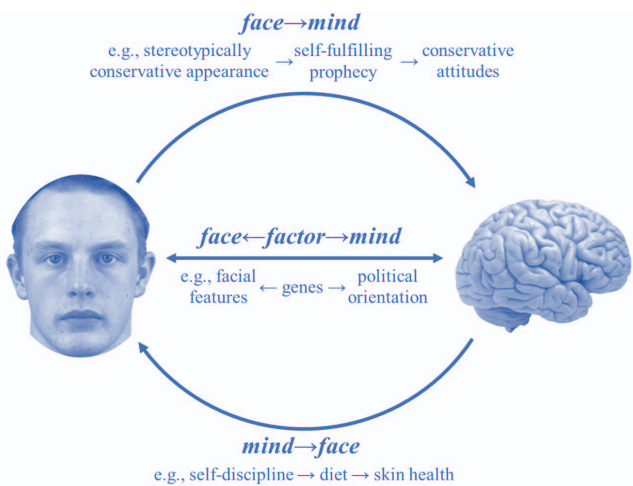

*Note.* The facial image was used with permission.

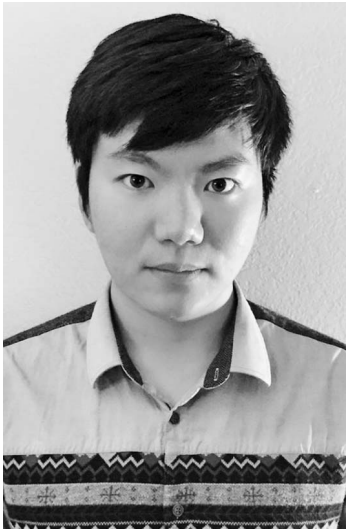

**Yilun Wang**

facial appearance and political orientation generalize beyond the carefully standardized facial images and the sample used in Studies 1 and 2. Finally, Study 4 explores the links between political orientation and several interpretable facial features.

Our studies were approved by the Stanford University institutional review board. We have complied with American Psychological Association ethical standards in the treatment of our participants. Informed consent was obtained from all participants. Informed consent for the publication of identifying facial images was obtained from the participants whose images were used in the figures. Politicians' images are in the public domain.

## Study 1: Standardized Images of Neutral Faces

We first show that political orientation can be algorithmically inferred from carefully standardized images of neutral faces while controlling for demographics, self-presentation, facial expression, and image properties.

We employ a commonly used facial recognition algorithm, VGGFace2, in ResNet-50-256D architecture (Cao et al., 2018). The neural network underlying this model was trained on 9,131 people and their 3.3 million facial images, varying in pose, facial expression, illumination, and other superficial characteristics. It was trained to convert diverse facial images of an individual person into *face descriptors*, or a numerical vector that is both unique to that individual and consistent across their different images. As a result, face descriptors tend to subsume distinctive and stable facial features. They are not readily interpretable: A single number might subsume several facial features that humans consider to be distinct (e.g., skin tone, facial width, and eye color).

Face descriptors are typically used to algorithmically recognize faces. Descriptors extracted from a given image are compared to those stored in a database. If they are similar enough, the faces are considered a match. Here, we use a linear regression to map face descriptors on a political orientation scale and then use this mapping to predict political orientation for a previously unseen face. In this way, we quantify the strength of the link between political orientation and facial characteristics encoded in face descriptors. The risk of overfitting (i.e., discovering links between political orientation and facial features that are specific to our sample) is reduced both by cross-validation and by the fact that VGGFace2 was trained on an independent sample and for a different purpose.

### Transparency and Openness

Author notes, code, and data used in the present study (excluding participants' pictures to protect their privacy) are available at https://osf.io/nuz2m (Kosinski et al., 2023). All studies employ secondary data. Data employed in Studies 1, 2, and 4 were collected based on a preregistration protocol (https://aspredicted.org/9mi5y.pdf). Study 2 was preregistered (https://aspredicted.org/py5vm.pdf). For all experiments, we have reported all measures, conditions, and data exclusions.

### Method

Study 1 employs secondary data collected in the course of a different study.

#### Participants

Participants ($n = 596$) were recruited at a major private university in exchange for financial compensation ranging from $20 at the beginning of the study to $40 at the end. To encourage honest and thoughtful responses, they were offered free feedback on their personality. Participants first filled out a battery of questionnaires and were then photographed. Data were collected between 2018 and 2019.

Participants self-reported their gender (57% female) and age ($M_{age} = 22$; interquartile distance = [19, 22]). None reported having undergone plastic surgery. We included 591 participants for whom age, gender, and score on the political orientation scale were available. The power analysis indicated that this secondary sample was sufficiently large to detect effects of $r = .12$ with a power of .8 at the significance level of .05.

#### Standardized Facial Images

The images were collected using a procedure designed in consultation with a professional photographer. Participants wore a black T-shirt adjusted using binder clips to cover their clothes. They removed all jewelry and—if necessary—shaved facial hair. Face wipes were used to remove cosmetics until no residues were detected on a fresh wipe. Their hair was pulled back using hair ties, hair pins, and a headband while taking care to avoid flyaway hairs. Participants sat up straight with their lower back pressed against the chair's back, their upper back off the chair, feet flat on the floor, and hands on the lap.

We used a neutral background. The Nikon D3200 camera (AF-S NIKKOR 35 mm 1:1.8 G lens) was positioned four and a half feet from the back of the chair and aligned with the middle of the participants' eyes.

To ascertain that participants were looking directly at the camera, we checked if the ears appeared to be of the same size and if the chin was facing the camera directly. Participants were instructed to:

> Relax your shoulders. Look toward the camera, and make sure your chin is at a 90-degree angle to your body. Finally, take a deep breath in (wait two full "Mississippi" seconds), and as you slowly exhale, relax all your facial muscles, and allow your facial expression to be as neutral as possible. Take another deep breath in (wait two full seconds) and slowly exhale. Please close your mouth and look directly at the camera.

We took multiple pictures per participant in 4,000 × 4,000 pixels resolution. A hypothesis-blind research assistant selected the best picture and cropped it in a tight rectangle from the top of the forehead (the start of the hairline) to the bottom of the chin, with the ears included. Cropped images were downsized to 224 × 224 pixels (see Figure 2). VGGFace2 in ResNet-50-256D architecture (Cao et al., 2018) was used to convert facial images into 256-value-long face descriptors.

### Ethnicity

Participants' perceived ethnicity ("Caucasian" vs. "non-Caucasian") was approximated by three hypothesis-blind research assistants who independently reviewed all facial images; 74% ($n = 436$) of participants were labeled as "Caucasian" by all three research assistants. The percentage of Caucasian participants was close to that of the U.S. population (76%; U.S. Census Bureau, 2020). The sample size was inadequate to conduct separate analyses for the 155 non-Caucasian participants, especially given that this subset included a diverse set of ethnicities.

### Political Orientation Scale

Political orientation was measured using five Likert-style items, listed in Table 1. Items 1 and 2 aimed at voting behavior; Items 3, 4, and 5 aimed at general, social, and economical political attitudes, respectively. Items pointing in the conservative direction were reversed, so higher scores represent liberal orientation. Responses to individual questions were normalized (to have a range from 0 to 1) and averaged to compute participants' political liberalism score. The scale was highly reliable (Cronbach's $\alpha = .94$) and closely aligned with participants' political party preferences (see Supplemental Materials for more details). Participants tended to be liberal (skew = −1.09). Age, Pearson product–moment correlation $r(589) = −.01$; $p = .85$; 95% CI [−.09, .07], and ethnicity, $r(589) = −.06$; $p = .12$; 95% CI [−.14, .02], were not significantly correlated with political orientation in our sample. Women tended to be more liberal, $r(589) = .24$; $p \leq .001$; 95% CI [.17, .32].

**Figure 2**
*Standardized Facial Image (Used With Permission)*

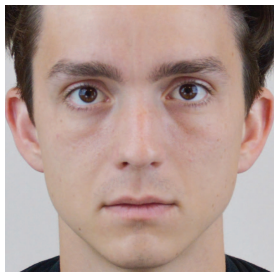

### Regression

Predictions of political orientation were produced using leave-one-out cross-validation: Each participant's score was predicted using a model derived from all other participants (i.e., training set) to ascertain that all predictions were made by a model that had not seen a given participant before. Independent variables—including age, gender, perceived ethnicity, or facial descriptors—were entered into a linear regression to predict participants' political orientation. Least absolute shrinkage and selection operator variable selection was used for facial descriptors: Parameter α was set to 1, and parameter δ was fitted separately for each training set using leave-one-out cross-validation. For ease of interpretation, predicted variables were always standardized (i.e., their mean was set to 0 and SD to 1). To avoid the leakage of information, the standardization was conducted separately within each training set.

### Results

The results are presented in Table 2. We first regress political orientation on age, gender, and ethnicity (Model 1). Only the coefficient for gender was statistically significant ($\beta = .5$; standard error [$SE$] = .08; $p < .001$). The resulting cross-validated prediction accuracy, expressed as the Pearson product–moment correlation between predicted and self-reported political orientation, equaled $r(589) = .23$ (95% CI [.15, .30]; $p < .001$). This is significantly above the expected correlation of $r = 0$ if there was no link between political orientation and these variables. The root-mean-square error (RMSE), or a difference between predicted and observed political orientation, equaled RMSE = .973. Observed RMSE can be compared against the baseline RMSE or RMSE of a baseline (null) model, where predicted scores are set to mean scores in the training set (RMSE$_{mean}$ = 1).

Predictive power offered by age, gender, and ethnicity provides a useful reference point for the accuracy of face-based models. Additionally, the residual of this model—or the political orientation score decorrelated with age, gender, and ethnicity—will be used to estimate the accuracy of face-based models while controlling for these variables. The correlation between the raw political orientation scores and the scores decorrelated with age, gender, and ethnicity equaled $r(589) = .97$ (95% CI [.96, .97]; $p \leq .001$). As expected, the correlation

**Table 1**
*Correlation Table for Political Orientation Scale Items (Before They Were Reversed and Normalized)*

| | | | | Pearson's $r$ | | | | |
|---|---|---|---|---|---|---|---|---|
| Item | M | SD | NAs | 1 | 2 | 3 | 4 | 5 |
| 1. I tend to vote for liberal political candidates. | 4.07 | 1.13 | 0 | — | | | | |
| 2. I tend to vote for conservative political candidates. | 1.82 | 1.13 | 0 | −.92 | — | | | |
| 3. To what extent do you consider yourself to be liberal or conservative in general? | 2.65 | 1.34 | 5 | −.87 | .87 | — | | |
| 4. To what extent do you consider yourself to be socially liberal or conservative? | 2.08 | 1.22 | 2 | −.74 | .74 | .79 | — | |
| 5. To what extent do you consider yourself to be economically liberal or conservative? | 3.34 | 1.60 | 2 | −.71 | .71 | .79 | .57 | — |
| Political orientation scale | 0 | 1 | 0 | .94 | −.94 | −.95 | −.83 | −.84 |

*Note.* NAs = number of missing values. All correlations are significant at $p \leq .001$. Items 1 and 2 had five response options: *strongly agree* to *strongly disagree*. Items 3, 4, and 5 had seven response options: *extremely liberal* to *extremely conservative*.

between decorrelated scores and age, gender, and ethnicity all equaled $r(589) = 0$ (95% CI [−08, .01]; $p = 1$).

Second, we regress political orientation scores on VGGFace2 face descriptors (Model 2). Out of 256 face descriptors, 65 had nonzero regression coefficients. The model's cross-validated prediction accuracy, $r(589) = .21$; 95% CI [.13, .28]; $p < .001$; RMSE = .988, was comparable with one observed for age, gender, and ethnicity. That result should not be surprising: Political orientation is linked with age, gender, and ethnicity, and these traits are clearly displayed on people's faces. (In our sample, only gender predicted political orientation.) It is also consistent with the results observed in a sample of nonstandardized social media profile images of one million participants (Kosinski, 2021), where the accuracy equaled $r = .38$ (derived from the area under the curve coefficient of 72%). The lower accuracy observed here could be driven by a much smaller training sample and by our efforts aimed at minimizing the role of self-presentation.

Third, we test the predictability of political orientation while controlling for age, gender, and ethnicity (Model 3). We regress the political orientation score decorrelated with these variables (or Model 1's residuals) on VGGFace2 face descriptors. The number of face descriptors employed by the model (46) was smaller than the one observed in Model 2. This is to be expected, as Model 3 could not benefit from information about participants' demographics. Yet, the model's resulting cross-validated accuracy equaled $r(589) = .22$ (95% CI [.15, .30]; $p < .001$; RMSE = .976), indicating that political orientation can be derived from facial images, even when controlling for self-presentation and demographics. To put Model 3's accuracy in perspective, consider the following widely known effects: Job interviews predict job success with an accuracy of $r = .20$, while alcohol intoxication increases aggressiveness by $r = .23$ (from Meyer et al., 2001). In other words, a single standardized image of a neutral face reveals political orientation about as much as job interviews reveal job success or alcohol drives aggressiveness. Moreover,

**Table 2**
*Regression Models and Their Cross-Validated Predictive Accuracy*

| | Regression coefficients | | | Models' cross-validated accuracy | | | | |
|---|---|---|---|---|---|---|---|---|
| Variable | Estimate | SE | p | n | r | 95% CI | p | RMSE |
| Model 1: Predicting political orientation | | | | | | | | |
| Intercept | −.12 | .18 | .49 | 591 | .23 | [.15, .30] | .001 | .973 |
| Gender | .50 | .08 | <.001 | | | | | |
| Age | .00 | .01 | .85 | | | | | |
| Caucasian | −.18 | .09 | .06 | | | | | |
| Model 2: Predicting political orientation | | | | | | | | |
| Face descriptors ($k = 256$) | Nonzero coefficients: 65 | | | 591 | .21 | [.13, .28] | <.001 | .988 |
| Model 3: Predicting political orientation decorrelated with age, gender, and ethnicity | | | | | | | | |
| Face descriptors ($k = 256$) | Nonzero coefficients: 46 | | | 591 | .22 | [.15, .30] | <.001 | .976 |
| Model 4: Predicting political orientation | | | | | | | | |
| Intercept | −.15 | .18 | .40 | 591 | .31 | [.24, .39] | <.001 | .950 |
| Gender | .51 | .08 | .001 | | | | | |
| Age | .00 | .01 | .97 | | | | | |
| Caucasian | −.20 | .09 | .03 | | | | | |
| Predicted values from Model 3 | .79 | .14 | .001 | | | | | |

*Note.* Regression parameters were estimated on the entire sample; separate models were fitted in each cross-validation fold. CI = confidence interval; SE = standard error; RMSE = root-mean-square error.

Model 3's accuracy was as high as the accuracy afforded by gender (Model 1), a demographic trait with widely known links with political orientation (Pew Research Center, 2018).

Fourth, we further test the additive value of face descriptors over demographic traits (Model 4). We regress political orientation on age, gender, ethnicity, and political orientation scores derived from face vectors while controlling for these variables. (The latter variable is a cross-validated output of Model 3.) As in Model 1, political orientation was predicted by gender ($\beta = .51$; $SE = .08$; $p < .001$). Even stronger predictive power was provided by face-derived political orientation scores ($\beta = .79$; $SE = .14$; $p < .001$). Model 4's overall cross-validated accuracy equaled $r(589) = .31$ (95% CI [.24, .39]; $p < .001$; RMSE = .950), significantly above the accuracy of the model based solely on age, gender, and ethnicity, Hotelling's $t(588) = 2.95$, $p = .003$.

One of the weaknesses of our approach is the simplistic treatment of participants' ethnicity, represented by a single Caucasian versus non-Caucasian variable. The 155 non-Caucasian participants represent a diverse set of ethnicities, but their number was inadequate to conduct meaningful statistical analyses. To partially address this issue and to further control for the effect of ethnicity, we replicate the analyses presented above while limiting the sample to Caucasian participants. The results presented in Supplemental Table S1 are very similar to one observed on the entire sample. For example, Model 4 A testing the additive value of face descriptors over demographic traits achieved an accuracy of $r(434) = .3$ (95% CI [.21, .38]; $p < .001$; RMSE = .954).

## Study 2: Human Raters

Study 1 indicates that facial recognition algorithms can predict participants' political orientation from carefully standardized facial images while controlling for demographics, self-presentation, facial expression, and image properties. Here, we replicate those results while replacing the algorithm with human raters.

### Method

Study 2 was preregistered (https://aspredicted.org/py5vm.pdf).

### Human Raters

Raters recruited on Amazon's Mechanical Turk ($n = 1,188$; location: United States only; 97% or higher acceptance rate) were each presented with five randomly selected facial images collected in Study 1. They were asked to rate the person in the picture using the same five political orientation scale items as in Study 1, as well as to report the person's gender. Raters produced 5,894 response sets. Following the preregistration protocol, we excluded 15 incomplete responses and 162 raters who misjudged the gender of two or more participants. The final sample contained 5,077 responses of 1,026 raters (8.6 ratings per face on average). The instructions presented to the raters are presented in Supplemental Materials.

### Raters' Perceptions

Raters' perceptions were scored in the same way as participants' self-reports in Study 1. The correlations between raters' responses to individual items are presented in Table 3. We note that Item 1, pointing in the liberal direction, correlates positively (but relatively weakly) with Items 2–5 that point in the conservative direction. The expected direction of those correlations is negative. This pattern shows that the raters tended to respond to all questions in the same manner, to some extent, irrespective of the content. Such a response bias typically arises from inattention or exceedingly challenging questions. Both factors are surely at play here: The human raters likely suffered from limited attention and struggled to predict political orientation based solely on a facial image. The exclusion of this question from the scale reduces its correlation with self-reported political

**Table 3**
*Correlation Table for Human Raters' Responses to the Perceived Political Orientation Scale Before They Were Reversed and Normalized*

| | | | Pearson's $r$ | | | | |
|---|---|---|---|---|---|---|---|
| Item | M | SD | 1 | 2 | 3 | 4 | 5 |
| 1. This person tends to vote for liberal political candidates. | 3.70 | 0.95 | — | | | | |
| 2. This person tends to vote for conservative political candidates. | 3.65 | 1.03 | .05 | — | | | |
| 3. To what extent does this person consider themselves to be liberal or conservative in general? | 4.32 | 1.96 | .12 | .48 | — | | |
| 4. To what extent does this person consider themselves to be socially liberal or conservative? | 4.31 | 1.96 | .09 | .49 | .60 | — | |
| 5. To what extent does this person consider themselves to be economically liberal or conservative? | 4.32 | 1.93 | .16 | .39 | .67 | .59 | — |
| Perceived political orientation scale | 0 | 1 | .10 | .69 | .83 | .82 | .80 |

*Note.* All correlations are significant at $p \leq .001$. Items 1 and 2 had five response options: *strongly agree* to *strongly disagree*. Items 3, 4, and 5 had seven response options: *extremely liberal* to *extremely conservative*.

orientation, showing that it contains a useful signal. The overall Cronbach's α reliability of this scale equaled .7.

## Results

We first checked to what extent raters' perceptions have been affected by photographed individuals' demographic traits. We regressed the perceived political orientation with photographed individuals' age, gender, and ethnicity. Only perceived ethnicity was significantly, but very weakly, linked with raters' perceptions (β = −.06; $p$ = .05). The cross-validated predictive accuracy of this model equaled $r(5,075)$ = .03 (95% CI [.00, .06]; $p$ = .02). The residual of this model—or raters' perceptions decorrelated with age, gender, and ethnicity—was employed in the next analysis. It correlated very highly with raw perception scores, $r(5,075)$ = .99; 95% CI [.00, .06]; $p$ = .02.

Next, we tested the perceptions' accuracy. As several raters rated each of the images, the perceptions (decorrelated with age, gender, and ethnicity) of a randomly selected subset of one to 15 raters were averaged to compute the aggregate perceived scores. The accuracy was estimated by correlating these aggregated judgments with self-reported scores (decorrelated with age, gender, and ethnicity) estimated in Study 1 (i.e., residuals of Model 1). To reduce the noise stemming from the random selection of raters, the process was repeated 1,000 times for each number of ratings per image. The resulting correlation coefficients were averaged using Fisher's $z$-transformation.

The results are presented in Table 4. Given one rating per image, the accuracy was not significantly different from 0, $r(589)$ = .02; $p$ = .67; 95% CI [−.06, .10]. Prediction accuracy increased with the number of aggregated ratings, reaching significance at eight ratings per image, $r(373)$ = .10; $p$ = .05; 95% CI [.00, .20], and peaking at 11 raters per image, $r(143)$ = .21; $p$ = .01; 95% CI [.05, .36].

While the accuracy decreased for 12 and more raters per image, this does not indicate that adding more raters decreases accuracy. Instead, as the number of images dropped, the results became more unstable (e.g., there were only 16 images with 15 or more ratings). Virtually identical results were observed for the subset of Caucasian participants (see Supplemental Table S2) and when employing the raw perception scores (which is unsurprising, as these variables were virtually identical).

## Study 3: External Validation—Facial Images of Politicians

The results of Studies 1 and 2 indicated that both algorithms and human raters can predict political orientation from standardized facial images while controlling for self-presentation and demographics. Here, we apply the model derived from the standardized images in Study 1 to predict political party affiliation in a very different sample: self-selected profile images of 3,401 members of the lower and upper chambers of parliament from the United States (1981–2018; $n$ = 1,826), the United Kingdom (1997–2018; $n$ = 1,024), and Canada (2011–2018; $n$ = 551). The power analysis indicated that the secondary sample used in Studies 1, 2, and 3 was sufficiently large to detect effects of $r$ = .12 with a power of .8 at the significance level of .05.

## Method

### Politicians

Politicians' official profile images, date of birth, gender, and party affiliation were downloaded from http://everypolitician.org; all data recorded until September 2018 were used. The images were processed following the procedure described in Study 1. When more than one facial image was available, the most recent was used. Politicians ($n$ = 3,332) were categorized as *liberal* if their party affiliation was designated as "Democrat" ($n$ = 897), "Labor" ($n$ = 445), "Liberal" ($n$ = 187), "NDP" ($n$ = 103), "Liberal Democrat" ($n$ = 81), "Liberal Party of Canada" ($n$ = 18), or "Social Democratic and Labor Party" ($n$ = 5); and *conservative* if their affiliation was described as "Republican" ($n$ = 929), "Conservative" ($n$ = 629), or "Conservative Party of Canada" ($n$ = 38). Politicians with other affiliations ($n$ = 156) were excluded from the analysis. The power analysis indicated that this secondary sample was sufficiently large to detect effects of $r$ = .05 with a power of .8 at the significance level of .05.

### Facial Images

Face++ was used to predict politicians' ethnicity: Only the subsets of Black ($n$ = 204) and White ($n$ = 2,969) participants were numerous enough to enable a meaningful statistical

**Table 4**

*Accuracy of Human Ratings When Predicting Political Orientation From Standardized Images While Controlling for Age, Gender, and Ethnicity*

| Raters per image | Unique images | $r$ | 95% CI | $p$ |
|---|---|---|---|---|
| 1 | 591 | .02 | [−.06, .10] | .67 |
| 2 | 591 | .02 | [−.06, .10] | .62 |
| 3 | 589 | .02 | [−.06, .10] | .61 |
| 4 | 578 | .03 | [−.05, .12] | .42 |
| 5 | 549 | .05 | [−.04, .13] | .26 |
| 6 | 502 | .07 | [−.02, .15] | .14 |
| 7 | 449 | .08 | [−.01, .18] | .08 |
| 8 | 375 | .10 | [.00, .20] | .06 |
| 9 | 285 | .14 | [.02, .25] | .02 |
| 10 | 207 | .17 | [.03, .30] | .02 |
| 11 | 145 | .21 | [.05, .36] | .01 |
| 12 | 94 | .16 | [−.05, .35] | .13 |
| 13 | 59 | .02 | [−.23, .28] | .85 |
| 14 | 35 | .08 | [−.26, .40] | .64 |
| 15 | 16 | .00 | [−.50, .49] | 1.00 |

*Note.* CI = confidence interval.

analysis. Face++ was also used to detect smiles: Faces with the probability of smiling above 90% were categorized as smiling; those below 10% were categorized as nonsmiling. As in Study 1, models' performance is expressed as the Pearson product–moment correlation.

### Face-Based Political Orientation

We applied the most conservative of the models (Model 3A) trained in Study 1: the model regressing political orientation (decorrelated with age and gender) on face vectors on a subset of Caucasian participants. (Similar results were achieved for Model 3, regressing political orientation—decorrelated with age, gender, and ethnicity—on face vectors of all participants.)

### Results

The results are presented in Table 5. The accuracy is expressed as the point-biserial correlation between face-based predictions and the dichotomous political orientation variable. The results show that the model trained in Study 1 can predict politicians' political orientation across countries, genders, ethnicities, age groups, and for smiling and not-smiling faces, with a median accuracy of $r = .13$ (interquartile range: [.10, .15]). The accuracy on the entire sample equaled $r(3,399) = .13$ ($p \leq .001$; 95% CI [.10, .17]).

That political affiliation can be predicted from self-selected profile images is unsurprising and has been previously shown in samples of politicians (Joo et al., 2015) and nonpoliticians (Kosinski, 2021; Rasmussen et al., 2023). Yet, that the model derived from standardized images obtained in Study 1 worked in a sample of self-selected images is remarkable. This model had little chance to discover (and thus employ) the associations between political orientation and factors that were held constant or were controlled for, such as facial expressions, demographics, self-presentation, and the properties of the image. It was trained to focus on stable facial features, yet such features are, to some extent, distorted on nonstandardized profile images by the variance in head orientation, facial hair, facial expression, image properties, and so forth. In particular, the model worked for smiling faces. This indicates that either smiling does not overtly distort facial features employed by the model or that VGGFace2 was able to correct for such distortions, as it was originally trained to recognize faces regardless of the facial expression, angle of the camera, and so forth.

Moreover, the model worked across countries, genders, age groups, and ethnicities. This is remarkable, as it suggests that the links between facial features and political orientation discovered by the model in the relatively young and liberal sample of U.S. participants generalized well beyond the boundaries of this sample. Overall, our results suggest that the predictability of political orientation extends beyond the high-quality images of the relatively young and relatively liberal participants used in Studies 1 and 2.

### Study 4: The Associations Between Facial Appearance and Political Orientation

Studies 1–3 indicate that facial appearance is associated with political orientation—beyond what is revealed by self-presentation, demographics, and image properties. Here, we try to identify some of those associations using facial images from Study 1. As the number of non-Caucasian participants ($n = 157$) was too small to conduct meaningful analyses (especially given that this subset included a diverse set of ethnicities), the analyses presented below are limited to 436 participants labeled as "Caucasian" by all three research assistants. For clarity, we present the results along with the methods.

### Heat Map

We start by mapping facial areas employed by the prediction model. Facial images were divided into a matrix of $32 \times 32$ squares ($7 \times 7$ pixels). For each square, we masked it across all images and reran the procedure employed in Study 1 to predict participants' political orientation (decorrelated with age and gender). Next, we computed the average absolute difference between the original predictions and those extracted from images after masking a given square. In other words, we manipulated the availability of information in a particular square and measured how it affected the model's predictions.

**Table 5**
*The Performance of Model 3A Trained on Standardized Images in Study 1 When Applied to Predict Political Orientation in a Sample of Naturalistic Facial Images of Politicians*

| Subsample | $n$ | $r$ | $p$ | 95% CI |
|---|---|---|---|---|
| All politicians | 3,401 | .13 | <.001 | [.10, .17] |
| Country | | | | |
|   United States | 1,826 | .15 | <.001 | [.11, .20] |
|   United Kingdom | 1,024 | .08 | .03 | [.02, .14] |
|   Canada | 551 | .15 | <.001 | [.06, .23] |
| Gender | | | | |
|   Female | 678 | .12 | .001 | [.05, .20] |
|   Male | 2,721 | .10 | <.001 | [.06, .14] |
| Smile | | | | |
|   Yes | 2,545 | .14 | <.001 | [.11, .18] |
|   No | 486 | .10 | .03 | [.01, .19] |
| Ethnicity | | | | |
|   Black | 204 | .26 | <.001 | [.13, .39] |
|   White | 3,037 | .12 | <.001 | [.09, .16] |
| Born in | | | | |
|   1900–1940 | 729 | .10 | .01 | [.03, .17] |
|   1941–1950 | 669 | .18 | <.001 | [.10, .25] |
|   1951–1958 | 678 | .10 | <.01 | [.02, .17] |
|   1959–1967 | 639 | .18 | <.001 | [.10, .25] |
|   1968–1994 | 627 | .15 | <.001 | [.07, .22] |

*Note.* CI = confidence interval.

The heat map presented in Figure 3 reveals that the facial areas employed by the model included the philtrum, eyebrows and eyes, nasal bridge, and mouth corners. Importantly, the background and hair played virtually no role in prediction, confirming that those factors were successfully controlled for.

### Average Facial Outlines

Next, we visually inspect the differences between average facial outlines and average faces of the most liberal and most conservative males and females (the top and bottom quartiles of the political orientation scale decorrelated with age and gender). The outlines were produced by averaging facial landmarks extracted using the Face++ algorithm (see Supplemental Figure S2). The average faces were generated by averaging the pixel values of facial images. To increase average faces' sharpness, images were first aligned along facial landmarks using a piecewise linear 2D transformation. To reduce the role of facial marks and facial asymmetry, horizontally flipped copies of each image were also included.

The results are presented in Figure 4. The average facial outlines (left column) suggest that liberals had smaller lower faces. This is also visible on the average faces (right column): Note that liberals' lips and noses are shifted downward, and their chins are smaller. Otherwise, the average outlines and faces are virtually identical, revealing no other obvious differences between liberals and conservatives, including in facial expression, grooming, skin color, or head orientation.

**Figure 3**
*Heat Map Representing the Average Absolute Change in the Standardized Prediction Score Resulting From Masking a Given Image Area*

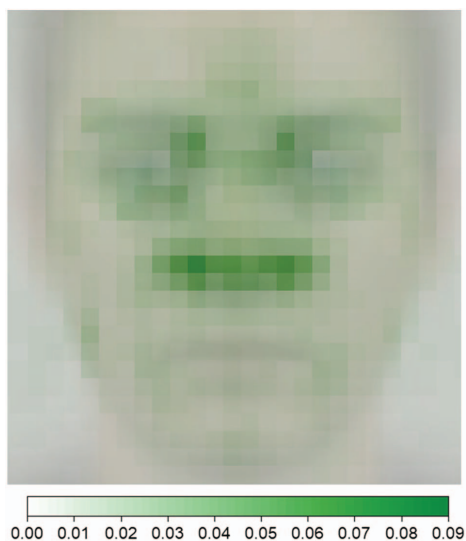

*Note.* The color scale ranges from transparent (no change) to dark green (an average absolute predicted score change of .09 *SD*).

### Facial Dimensions

We further examine the link between political orientation and facial morphology revealed by the facial outlines. We followed an established procedure (Kosinski, 2017) to automatically extract facial width, upper and lower facial height, pupillary distance, and philtrum from FPP++ facial landmarks (Supplemental Figure S2 visually explains those dimensions). Those dimensions were used to compute two widely used facial measures: facial width-to-height ratio (fWHR; $\frac{\text{facial width}}{\text{upper facial height} + \text{philtrum}}$) and lower face size $\left(\frac{\text{lower facial height}}{\text{pupillary distance}}\right)$. Additionally, body mass index (BMI) was computed for 274 participants who self-reported their weight and height.

Table 6 presents the correlations between those variables and political orientation (raw scores; scores decorrelated with age and gender). The raw political orientation scores correlated with all variables apart from age and fWHR. Yet, these correlations could be largely attributed to the effect of gender: Political orientation was decorrelated with age, and gender correlated only with lower face size; liberals tended to have smaller faces; $r(434) = -.14$; $p = .003$; 95% CI [−.23, −.05].

We also built a linear regression model predicting political orientation (decorrelated with age and gender) from the lower face size, BMI, fWHR, and weight (as well as the interactions between those variables and gender). Only the lower face size was a significant predictor ($\beta = -.15$; $p = .028$). This confirms the relative importance of the lower face size and suggests that the link between those variables and political orientation is not mediated by gender.

How does the predictive power of the lower face size and BMI compare with the predictive power of the facial recognition algorithm estimated in Study 1? Would the VGGFace2-based model trained in Study 1 perform better if it was supplemented with explicit measures of lower face size and BMI? To answer these questions, we trained a series of regression models predicting political orientation (while controlling for age and gender) and used leave-one-out cross-validation to estimate prediction performance.

The predictive power of the lower face size equaled $r(434) = .11$; $p = .02$; 95% CI [.01, .20]. BMI's predictive power was insignificant $r(272) = .06$; $p = .36$; 95% CI [−.06, .18]. Combining the VGGFace2-based predictions (estimated in Study 1) with BMI, lower face size, and with both these variables did not improve prediction performance. The highest performance was afforded by combining VGGFace2 predictions with lower face size. Yet, this model's performance, $r(434) = .21$; $p < .001$; 95% CI [.12, .30], was no higher than the performance of the VGGFace2 predictions alone, $r(434) = .22$; see Study 1.

### Average Faces

Visual inspection of average faces did not reveal any clear differences in skin color or facial expression, which suggests that we appear to have successfully controlled for those variables. We further verified the lack of association between eye and skin color and political orientation by examining individual images. The average colors (operationalized as lightness, red, green, and blue) were extracted from a 40 × 20 pixels fragment of each pupil and an 80 × 80 pixels fragment of each upper cheek taken from standardized facial images in full resolution (4,000 × 4,000 pixels). None of those variables correlated significantly with political orientation. Even when entered into a leave-one-out cross-validated linear regression, their combined predictive power was not better than chance, $r(434) = .02$; $p = .61$; 95% CI [−.07, .12].

**Figure 4**
*Average Facial Outlines (Left Panel) and Average Faces (Right Panel) of Participants Occupying the Top and Bottom Quartiles on the Political Orientation Scale*

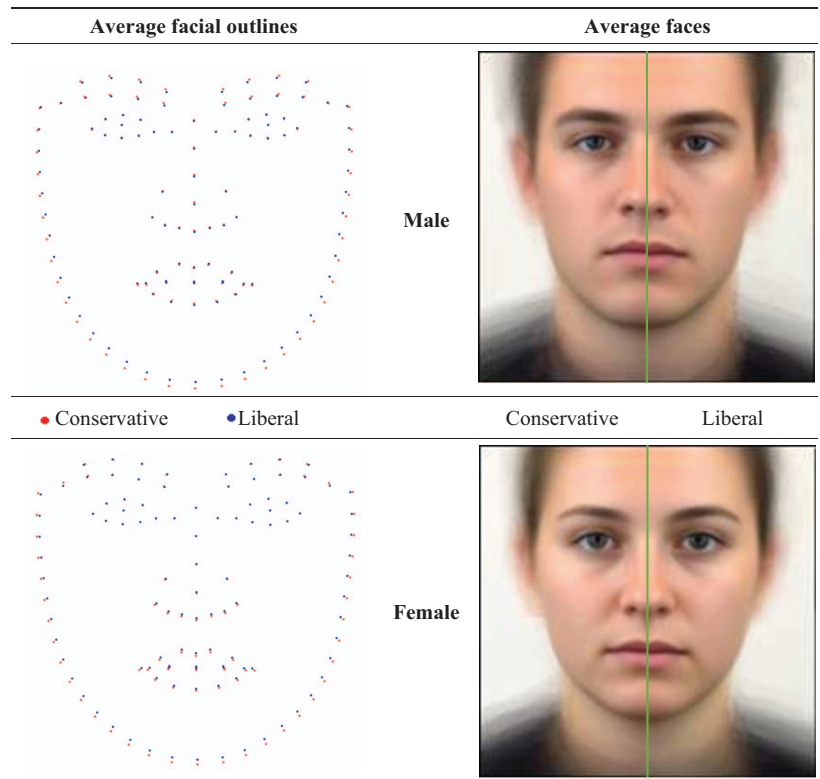

Similarly, the average faces maintain a neutral expression, and the expression does not differ between liberals and

**Table 6**
*Correlation Between Political Orientation and Interpretable Traits*

| Trait | M | SD | n | Pearson's r | | | | | | | |
|---|---|---|---|---|---|---|---|---|---|---|---|
| | | | | 1 | 2 | 3 | 4 | 5 | 6 | 7 | 8 |
| 1. Political orientation (raw) | 0.00 | 1.00 | 436 | — | | | | | | | |
| 2. Political orientation | 0.00 | 1.00 | 436 | .97*** | — | | | | | | |
| 3. Gender | 0.59 | 0.49 | 436 | .25*** | .00 | — | | | | | |
| 4. Age | 22.09 | 5.64 | 436 | −.03 | .00 | .06 | — | | | | |
| 5. fWHR | 2.06 | 0.37 | 436 | .03 | .04 | −.03 | −.04 | — | | | |
| 6. Lower face size | 1.02 | 0.09 | 436 | −.26*** | −.14** | −.47*** | .12** | .02 | — | | |
| 7. BMI | 23.14 | 2.90 | 274 | −.16** | −.12 | −.16** | .08 | .14* | .28*** | — | |
| 8. Weight (kg) | 69.51 | 12.36 | 274 | −.23*** | −.08 | −.56*** | .00 | .17** | .43*** | .76*** | — |
| 9. Height (cm) | 172.87 | 9.82 | 280 | −.21*** | −.03 | −.69*** | −.08 | .11 | .37*** | .09 | .71*** |

*Note.* Results estimated on 436 participants labeled as "Caucasian" by all three research assistants. n = number of available values. fWHR = facial width-to-height ratio; BMI = body mass index.
* $p \leq .05$. ** $p \leq .01$. *** $p \leq .001$.

conservatives. Yet, participants' faces may have still contained emotion-resembling cues, microexpressions, or traces of recently expressed emotions (Adams et al., 2012). To investigate the role of such factors, the Face++ algorithm was used to estimate facial expressions (sadness, disgust, anger, surprise, fear, and happiness; see Küntzler, Höfling & Alpers, 2021, for the discussion of the accuracy of automated judgments). None of these ratings correlated significantly with political orientation. Even when entered into a leave-one-out cross-validated linear regression, their combined predictive power was not better than chance, $r(434) = .00$; $p = .999$; 95% CI [−.09, .09].

## Discussion

Our results suggest that carefully standardized images of neutral faces reveal participants' political orientation, even when controlling for age, gender, ethnicity, self-presentation, and image properties. Both facial recognition algorithms ($r = .22$) and human raters ($r = .21$) could predict participants' political orientation, on par with how well job interviews predict job success, or alcohol drives aggressiveness (from Meyer et al., 2001), and comparable with the average effect size reported in the social and personality psychology literature ($r = .19$; interquartile range = [.11, .29]; Gignac & Szodorai, 2016). The effect was even higher ($r = .31$) when we leveraged face-derived information about participants' age, gender, and ethnicity. Moreover, the model derived from standardized images could successfully predict political orientation from the profile images of politicians. The performance ($r \approx .13$) was held across countries (the United States, the United Kingdom, and Canada), genders, ethnicities, age groups, and for smiling and nonsmiling faces, suggesting that the associations between facial appearance and political orientation generalize beyond the relatively young and liberal sample of the U.S. participants collected in this study.

The estimates of the effect size are likely conservative. First, the sample used in Study 1 was relatively young, and thus some of the potential mechanisms linking political orientation and facial appearance (discussed in the Introduction) had little time to take effect. For instance, the potential *self-discipline* → *diet* → *skin health* pathway is unlikely to have left many traces on the faces of participants with a median age of 22. Second, the low-resolution (255 × 255 pixels) two-dimensional facial images used here are an imperfect representation of facial appearance. High-resolution three-dimensional facial scans would likely convey more relevant signals. Third, a larger sample would likely enable the algorithm to discover and use more relevant signals. Consider, for example, that the accuracy observed on a sample of one million nonstandardized facial images reached $r = .38$ (Kosinski, 2021). Fourth, the VGGFace2 algorithm used here was not trained specifically to predict political orientation but to recognize individuals across images. An algorithm trained specifically to predict political orientation, or a more modern algorithm, would likely perform better; after all, facial recognition technology is still in its infancy. Fifth, our participants were predominantly liberal, limiting the algorithm's ability to discover links between faces and political orientation and thus reducing its performance. Finally, our sample was too small to build gender-specific models. While the model's external validity was equal for both men and women (see Study 3), many of the links between the face and political orientation may be gender-specific. Additional insights (and better predictive accuracy) could be afforded by analyzing men and women separately.

While the results of any single study should be taken with caution, our findings suggest that stable facial features are associated with political orientation. This should not be a controversial conclusion. As discussed in the introduction, given the number and diversity of the well-documented mechanisms implying that such associations should exist, it would be extraordinary if they did not. Acknowledging their existence—and studying them—could boost our understanding of political orientation's origins and consequences, the social and interpersonal consequences of facial appearance, and the effects of ideology on appearance. For example, many of the factors that can be gleaned from faces—such as endocrinological, genetic, or developmental factors—are challenging to measure in other ways. Thus, exploring the associations between political orientation and facial features could lead to novel insights into the role of hormones, genes, and the developmental environment in the formation of political orientation.

Among the interpretable facial features examined here, only lower face size was clearly associated with political orientation (liberals tended to have smaller lower faces); consistent with previous findings, fWHR showed no predictive power (Kosinski, 2017). Yet, the scarcity of apparent differences between conservative and liberal faces is not necessarily surprising: If they existed, they would likely be well known already. Instead, the neural networks underlying judgments made by both algorithms and human brains likely employ patterns that are too subtle or intricate to be apparent or easily interpretable. This is not unusual and applies to other facial attributions as well. For example, people show much agreement when judging attractiveness, trustworthiness, or competence from others' faces, yet we still lack in our understanding of the exact facial features driving many of these judgments (Todorov et al., 2015).

This work has many limitations. First, while the ethnicity of participants in our main sample (Study 1) was representative of the U.S. population (U.S. Census Bureau, 2020), the subset of non-Caucasians was too small to analyze in more detail. Second, our study focused on Western societies. Future research should focus on more diverse samples and other regions. Third, while the liberal–conservative dimension well describes political attitudes across contemporary capitalist societies (38), future research should explore these effects using different measures of political ideology and in other sociopolitical contexts. Fourth, while we controlled for head orientation (both at the picture-taking stage and later, by averaging facial descriptors for horizontally flipped faces), it might have played some role in the prediction. It was unlikely to be substantial: Past analysis of self-selected naturalistic images indicated that even when head orientation is not controlled for, it is only weakly associated with political orientation ($r = .10$; Kosinski, 2021). The use of 3D facial scans would allow more control.

Perhaps most crucially, our findings suggest that widespread biometric surveillance technologies are more threatening than previously thought. Previous research showed that naturalistic facial images convey information about political orientation and other intimate traits (Kachur et al., 2020; Kosinski, 2021; Rasmussen et al., 2023; D. Wang, 2022; Y. Wang & Kosinski, 2018), but it was unclear whether the predictions were enabled by self-presentation, stable facial features, or both. Our results, suggesting that stable facial features convey a substantial amount of the signal, imply that individuals have less control over their privacy. The algorithm studied here, with a prediction accuracy of $r = .22$, does not allow conclusively determining one's political views, in the same way as job interviews, with a predictive accuracy of $r = .20$, cannot conclusively determine future job performance. Nevertheless, even moderately accurate algorithms can have a tremendous impact when applied to large populations in high-stakes contexts. For example, even crude estimates of people's character traits can significantly improve the efficiency of online mass persuasion campaigns (Kosinski et al., 2013; Matz et al., 2017). Scholars, the public, and policymakers should take notice and consider tightening policies regulating the recording and processing of facial images.